\documentclass[10pt,journal,compsoc]{IEEEtran}
\usepackage{multirow}
\usepackage{amsmath,amsfonts}
\usepackage{array}
\usepackage[caption=false,font=normalsize,labelfont=sf,textfont=sf]{subfig}
\usepackage{textcomp}
\usepackage{stfloats}
\usepackage{url}
\usepackage{verbatim}
\usepackage{graphicx}
\usepackage{cite}
\usepackage{makecell}
\usepackage[table,xcdraw]{xcolor}
\usepackage{float}
\usepackage{color,xcolor}
\usepackage{fancyhdr}
\usepackage{booktabs}
\usepackage{pifont}      
\usepackage{bbding}      
\usepackage{fontawesome}

\makeatletter
\newif\if@restonecol
\makeatother

\usepackage[ruled]{algorithm2e}
\usepackage{cite}
\ifCLASSINFOpdf
\else
\fi
\begin{document}
\title{PointSee: Image Enhances Point Cloud %: Point Semantic Feature Enhancement for 3D Object Detection
}

\author{Lipeng Gu, Xuefeng Yan, Peng Cui, Lina Gong, Haoran Xie, Fu Lee Wang, Jin Qin, and Mingqiang Wei
        % <-this % stops a space
\thanks{L. Gu, X. Yan, L. Gong and M. Wei are with School
of Computer Science and Technology, Nanjing University of Aeronautics
and Astronautics, Nanjing, China (e-mail: 
glp1224@163.com; yxf@nuaa.edu.cn; gonglina@nuaa.edu.cn; mingqiang.wei@gmail.com).}
\thanks{P. Cui is with Institute of Software and Simulation, Dalian Naval Academy, Dalian, China (e-mail: cphkzy@126.com).}
\thanks{H. Xie is with Department of Computing and Decision Sciences, Lingnan University, Hong Kong, China (e-mail: hrxie@ln.edu.hk).}
\thanks{F. L. Wang  is with School of Science and Technology, Hong Kong Metropolitan University, Hong Kong, China (e-mail: pwang@hkmu.edu.hk).}
\thanks{J. Qin is with School of Nursing, The Polytechnic University of Hong Kong, Hong Kong, China (e-mail: harry.qin@polyu.edu.hk).}
}


\markboth{Journal of \LaTeX\ Class Files,~Vol.~14, No.~8, August~2015}%
{Shell \MakeLowercase{\textit{et al.}}: Bare Demo of IEEEtran.cls for Computer Society Journals}



\IEEEtitleabstractindextext{
\begin{abstract}
There is a trend to fuse multi-modal information for 3D object detection (3OD). However, the challenging problems of low lightweightness, poor flexibility of plug-and-play, and inaccurate alignment of features are still not well-solved, when designing multi-modal fusion newtorks.
We propose \textbf{PointSee}, a lightweight, flexible and effective multi-modal fusion solution to facilitate various 3OD networks by \textbf{se}mantic feature \textbf{e}nhancement of LiDAR \textbf{point} clouds assembled with scene images. 
Beyond the existing wisdom of 3OD, PointSee consists of a hidden module (HM) and a seen module (SM): HM decorates LiDAR point clouds using 2D image information in an offline fusion manner, leading to minimal or even no adaptations of existing 3OD networks; SM further enriches the LiDAR point clouds by acquiring point-wise representative semantic features, leading to enhanced performance of existing 3OD networks.  
Besides the new architecture of PointSee, we propose a simple yet efficient training strategy, to ease the potential inaccurate regressions of 2D object detection networks. 
Extensive experiments on the popular outdoor/indoor benchmarks show numerical improvements of our PointSee over twenty-two state-of-the-arts.
\end{abstract}

}

\maketitle

\section{Introduction}
LiDAR and camera are two common yet different-type sensors mounted on autonomous driving vehicles (e.g., nuScenes self-driving cars
possess one LiDAR and six cameras \cite{nuscenes}) for environmental perception such as 3D object detection (3OD). For the two sensors, LiDARs capture the spatial geometry information of 3D scenes, which are represented by point clouds \cite{venet,ips300+};
cameras are often used to take photos (RGB images) of 3D scenes, which contain rich semantic information. Recent years have witnessed considerable attempts to fuse the two-modal information for 3OD, since the spatial structural features from LiDAR point clouds and the semantic features from camera images have the potential to complement each other to enhance the performance of cutting-edge 3OD networks \cite{cl3d,msl3d,lift,wang2022multi,autoalign,wang2021multi,maff}.  

\begin{figure}[ht]
    \centering
    \includegraphics[width=0.45\textwidth]{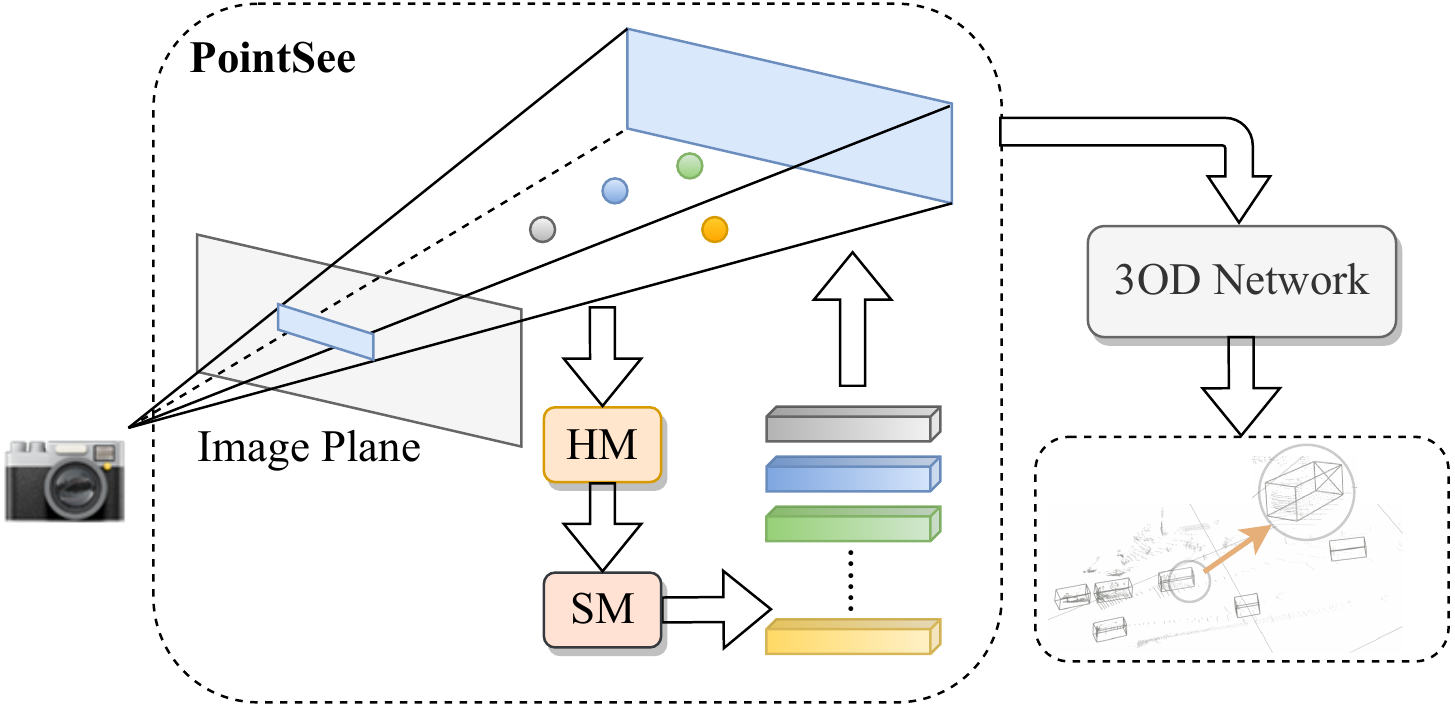}
    \caption{\textbf{Illustration of PointSee.} PointSee is a lightweight, plug-and-play and effective multi-modal fusion module to serve cutting-edge 3OD networks. We design a Hidden Module (HM) and a Seen Module (SM) in PointSee. HM behaves with an offline fusion manner to avoid any adaptation of existing 3OD networks, while SM brightens the ``eye" of 3OD networks by point-wisely appending additional semantic features to each point. In detail, we generate \textit{frustum point clouds} by 2D bounding boxes which are output by any 2OD network. Then, \textit{frustum point clouds} are fed into HM for recoding, which adds three kinds of label information for each point. After that, the recoded points are fed into  SM for obtaining representative semantic features, which is used to decorate LiDAR points. Finally, the decorated points can serve various 3OD networks.}
    \label{fig: pipeline}
\end{figure}

There are currently three types of multi-modal fusion solutions, i.e., point-level fusion, proposal-level fusion, and result-level fusion\cite{qian20223d,bai2022infrastructure}. First, the point-level fusions, such as MVX-Net \cite{MVX-Net}, EPNet \cite{EPNet}, PointAugmenting \cite{PointAugmenting} and PointPainting \cite{PointPainting}, utilize the mapping relationship between LiDAR points and pixels of camera images to enhance each LiDAR point with additional point-wise features from the images. However, these additional point-wise features (e.g., CNN features, segmentation scores) are difficult to acquire or to align with LiDAR points. 
Second, the proposal-level fusions, such as MV3D \cite{MV3D} and AVOD \cite{AVOD}, fuse the region proposals generated from both images and point clouds to produce higher-quality proposals, leading to heavy computations.
Third, the result-level fusions, such as F-PointNet \cite{F-PointNet} and F-ConvNet \cite{F-ConvNet}, utilize the outputs of 2D
object detection (2OD) networks to generate a series of \textit{frustum point clouds} as the region proposals and perform 3OD in each individual region proposal. This type of fusion regresses at most one object of interest in each 3D bounding frustum independently and is hard to apply into the off-the-shelf 3OD networks. Thus, these methods may fail once occlusion (i.e., two or more objects of interest in a \textit{frustum point cloud}) exists.
Moreover, most of existing fusion solutions have complex data processing pipelines or require large changes to the structure of off-the-shelf 3OD networks. 

We propose \textbf{PointSee}, a lightweight, flexible and effective multi-modal fusion solution to facilitate various 3OD networks by semantic feature enhancement of LiDAR point clouds assembled with scene images. PointSee mainly consists of a hidden module and a seen module (see Fig.~\ref{fig: pipeline}). 
 
\textit{Why is the hidden module hidden?} The hidden module recodes point clouds \textbf{offline} in order to obtain the label information required for subsequent modules. This offline operation avoids bringing any complex data processing pipeline into cutting-edge 3OD networks, thus enabling the lightweight and plug-and-play characteristics of PointSee. 
 
\textit{Why is the seen module seen?} The seen module point-wisely appends additional representative semantic features to the point cloud, making it possible to ``\textbf{brighten the eyes}'' of cutting-edge 3OD networks, so as to obtain better detection results. Such additional semantic features are obtained by fusing the high-dimensional features output by the 3D instance segmentation (3IS) network with the features of the point cloud itself (such as reflection intensity and RGB values) through multiple MLP layers. 
 
Our contributions are summarized as follows:
\begin{itemize}
\item We propose a \textit{light-weight} and \textit{plug-and-play} module \textbf{PointSee}, which consists of a hidden module (HM) and a seen module (SM), and can be plugged into off-the-shelf 3OD networks \textit{with any data augmentation} to improve their performance.
\item We propose a \textbf{hidden module} (HM) for offline recoding of point clouds to make PointSee more flexible with almost no network structural changes.
\item We propose a \textbf{seen module} (SM) for obtaining additional representative semantic features of the point cloud to obtain better detection results.
\item We extensively validate the effectiveness of PointSee via 3DSSD \cite{3DSSD} and PointRCNN \cite{PointRCNN} on the outdoor KITTI dataset \cite{kitti} and F-ConvNet \cite{F-ConvNet} on the indoor SUN-RGBD dataset \cite{sunrgbd}. The 3DSSD with PointSee outperforms state-of-the-art PV-based methods, such as PV-RCNN \cite{pv-rcnn}.
\end{itemize}

\begin{figure*}[ht]
    \centering
    \includegraphics[width=\textwidth]{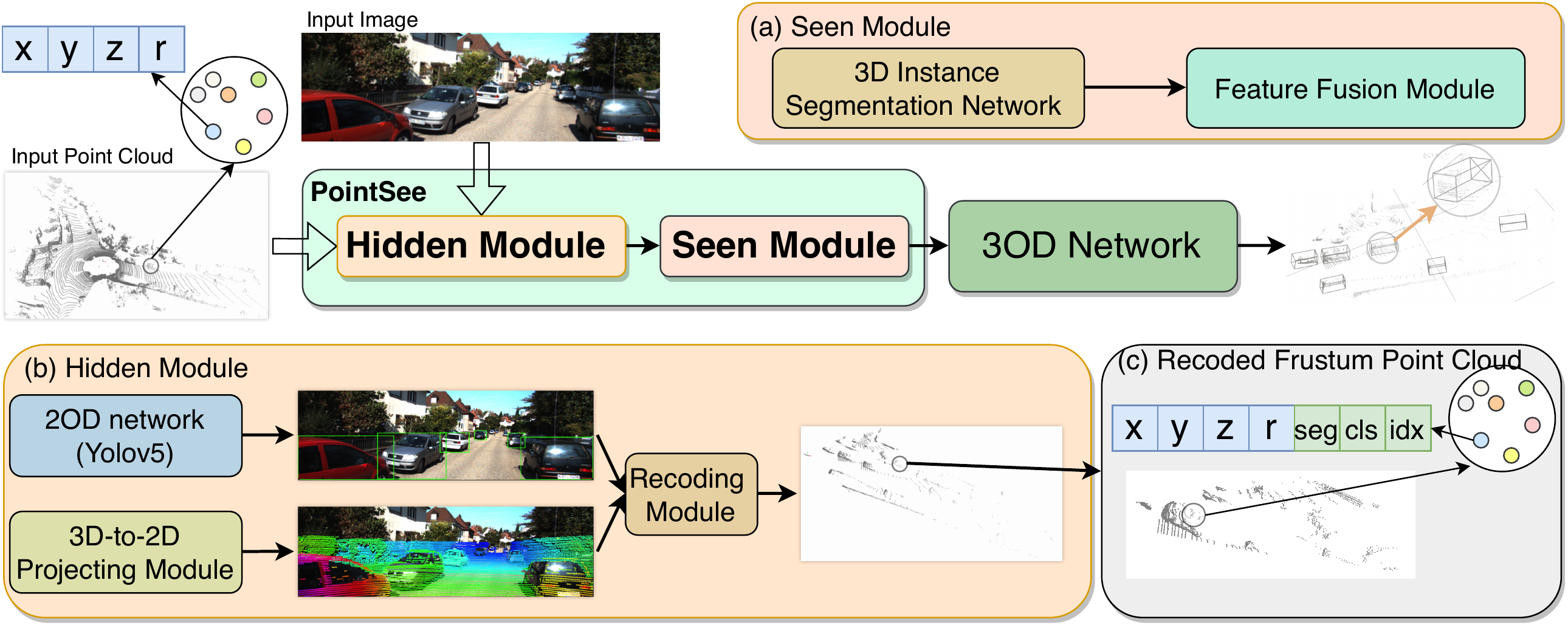}
    \caption{\textbf{Overview of PointSee.}  PointSee includes two parts: (1) Hidden Module; (2) Seen Module. The hidden module extracts \textit{frustum point clouds} based on 2D bounding boxes output by the 2OD network, and then expands three dimensions (\textit{semantic segmentation label}, \textit{category label} and \textit{object number}) for retained points. The seen module extracts additional point-wise semantic features to decorate these retained points for better detection results.}
    \label{fig:PointSee overview}
\end{figure*}

\section{Related Work}
We first introduce image-based 2D object detection and LiDAR-based 3D detection, and then introduce 3D object detection based on point-level, proposal-level and result-level methods in detail.
\subsection{Image-based 2D Detection.} 
After the development of the last decade, 2OD networks have achieved great success, especially in terms of high precision and real-time. 2OD networks are mainly divided into three categories: two-stage, single-stage and anchor-free networks. 
The two-stage network \cite{faster-rcnn, RCNN, sppnet, Fastrcnn, pyramid-networks}, also dubbed region proposal-based networks, first generates a large number of coarse region proposals and then further refines 2D bounding boxes on the generated region proposals, which causes the problem of high computational load. 
In order to improve detection efficiency, the single-stage network \cite{yolo, yolov4, SSD, Retina-Net} removes the process of region proposal generation and directly generates the category probabilities and location coordinates of objects of interest based on a series of anchors. Early single-stage networks traded detection accuracy for higher inference speed, while modern single-stage networks are surprising in inference speed and detection accuracy.
Unlike the single-stage network, the anchor-free network \cite{cornernet, centernet, fcos, ExtremeNet} directly perceives the region of objects of interest on the image and then yields the category probabilities and location coordinates.
In our PointSee, we use the latest 2OD network \textbf{Yolov5} to generate high quality 2D bounding boxes for \textit{frustum point clouds}.

\subsection{LiDAR-based 3D Detection.} 
Existing LiDAR-based methods only input the raw point cloud, and can be generally divided into three categories: voxel-based, point-based and PV-based. Voxel-based methods \cite{SECOND, VoxelNet, centerpoint, Part-A2} divide the point cloud into regular voxels and then use 3D CNNs to extract voxel features for 3OD. Point-based methods take the whole raw point cloud as input, extract a small number of key points and their features via some SA layers, and then carry out 3D detection. Point-based methods \cite{3DSSD, pointnet, IA-SSA, fast-pointrcnn, lidar-rcnn} are represented by the one-stage 3DSSD \cite{3DSSD} and the two-stage PointRCNN \cite{PointRCNN}. Different from PointRCNN, 3DSSD disposes all upsampling layers and refinement modules for computational efficiency. PV-based methods \cite{pv-rcnn} combine the advantages of point-based and voxel-based methods, but have large computation load. In this work, 3DSSD and PointRCNN are used as baselines for evaluating \textbf{PointSee}.

\subsection{Fusion-based 3D Detection.} 
Since images and point clouds provide complementary information, many fusion-based 3OD networks have made considerable achievements in recent years to obtain better 3D detection performance. Existing multi-modal fusion methods are generally classified into point-level, proposal-level and result-level methods. 
\textbf{Point-level fusion.} Point-level fusion methods \cite{MVX-Net,EPNet,PointAugmenting,PointPainting,epnet++} enhance the performance of 3OD by enriching the features of the point cloud. Concretely, these methods point-wisely decorate the point cloud using the features (e.g., CNN features, segmentation scores) extracted from the image. However, these point-level methods all suffer from the camera-LiDAR alignment problem, as concluded by DeepFusion \cite{deepfusion}, ``existence of the alignment problem when projecting LiDAR points into the image due to geometry-related data augmentation". In order to avoid this problem, our proposed PointSee simply omit the projecting process and directly obtain semantic features on the point cloud by the seen module.\\
\textbf{Proposal-level fusion.} Proposal-level methods \cite{MV3D,AVOD,AVOD-FPN,cm3d} first generate a large number of region proposals on the image and point cloud respectively, then fuse these region proposals and then further refines 3D bounding boxes based on fused region proposals. Due to the complex network structure, these methods fail to balance inference speed and detection performance.\\
\textbf{Result-level fusion.} Result-level methods \cite{F-PointNet,F-ConvNet,f-fusionnet} first employ 2D bounding boxes output by the 2OD network to generate a series of \textit{frustum point clouds}, and then perform 3OD based on each individual \textit{frustum point clouds}. These methods are network specific, not plug-and-play, fail in the combination of state-of-the-art 3OD networks. Different from the previous methods F-PointNets \cite{F-PointNet} and F-ConvNet \cite{F-ConvNet}, PointSee can be embedded in any off-the-shelf point-based 3D detection networks, and use additional semantic features to decorate raw point clouds for better detection results (see Table~\ref{tab1}).\\

\section{methodology}
We first introduce the motivation and overall structure of the proposed PointSee in Section 3.1, then introduce the hidden and seen modules in PointSee in Sections 3.2 and 3.3, respectively, followed by 3OD networks and datasets used to validate PointSee in Section 3.4, and finally the proposed training strategy in Section 3.5.
\subsection{Overview}
Existing multi-modal fusion efforts often fail to meet the requirements of lightweightness, flexibility of plug-and-play and accurate alignment of features.
For example, they may have complex data processing and fusion pipelines, leading to low lightweightness; they may introduce image features for certain 3OD networks and are difficult to transfer to other 3OD networks, leading to poor flexibility of plug-and-play; also, multiple LiDAR points may be projected to the same pixel in an image, leading to inaccurate alignment of point-to-pixel features. 

We introduce PointSee that is powerful and flexibly embedded in various 3OD networks. PointSee takes point clouds and images as input to enhance point semantic features for better performance of 3OD. 
As shown in Fig.~\ref{fig:PointSee overview}, PointSee consists of a hidden module and a seen module. 
The hidden module is an offline processing module, which is not required to embed in 3OD networks. Only the seen module is embedded in the head of 3OD networks for end-to-end training and inference.

\subsection{Hidden Module}
The hidden module (see Fig.~\ref{fig:PointSee overview} (b) adheres to the idea of ``label-following-point", which means that each point is appended with additional label information, so that the recoded point cloud is fed to the 3OD network without concerning the data processing and data augmentation pipelines. Concretely, the hidden module contains three parts: an image-based 2OD network, a 3D-to-2D projecting module, and a recoding module. 

\begin{algorithm}[!ht]  %其中这里面不能有H不然会报错，不过不影响结果
	\caption{Hidden Module}%算法名字
	\label{hidden module}
	\LinesNumbered %要求显示行号
	\KwIn{\\LiDAR points: $P \in \mathbb R ^{N,D}$, with $N$ points and $D \geq$ 3.
        \\Detections: $Det\_2D \in \mathbb R ^{M, 5}$, with $M$ detections.
        \\3D bounding boxes: $GT\_3D \in \mathbb R ^{K, 7}$, with $K$ labels.
        \\Homogenous transformation matrix: $T \in \mathbb R ^{4,4}$.\\ Camera matrix: $C \in \mathbb R ^{3,4}$.\\ \quad}
    \KwOut{Recoded LiDAR points: $P^{'} \in \mathbb R ^{N,D+3}$.\\ \quad} 

    $P^{'} \leftarrow Zero(0, D+3)$ \\
    $num \leftarrow 0$ \\
	\For{$det\_2d \in Det\_2D$}
	{
	    \For{$p \in P$}
	    {
	        $p_{image} \leftarrow PROJECT(C,T,p)$ \\
	        \If{$p_{image} \ \textbf{inside} \ det\_2d[:4]$}
	        {
	            $p \leftarrow Concatenate(p, 0)$ \\
	            \For{$gt\_3d \in GT\_3D$}
	            {
	                \If{$p \ \textbf{inside} \ gt\_3d$}
                    {
                        $p[-1] \leftarrow 1$ 
                    }
	            }
	            $p \leftarrow Concatenate(p, det\_2d[4])$ \\
	            $p \leftarrow Concatenate(p, num)$ \\
	            \textbf{add} $p$ to $P^{'}$ \\
	             $num \leftarrow num + 1$ \\
	        }
	    }
	}
	\Return $P^{'}$
\end{algorithm}

We use \textbf{Yolov5} as the image-based 2OD network to predict objects of interest in the image and output their 2D bounding boxes $Det\_2D=\left\{(x_1,y_1,x_2,y_2,c)\right\}_{i=0}^{M-1}$, where $x_1,y_1,x_2,y_2$ represent the coordinates of the upper-left and lower-right corners of the 2D bounding box, $c$ represents the category and $M$ represents the number for detected objects in a frame. Based on the known Homogenous transformation matrix $T$ and camera matrix $C$, the 3D-to-2D projecting module projects LiDAR points onto the image plane, and then captures all points \textit{inside} the 2D bounding boxes to form multiple \textit{frustum point clouds}. As shown in Fig.~\ref{fig:PointSee overview} (c), there is at least one object in each \textit{frustum point cloud}.

\begin{figure*}[ht]
    \centering
    \includegraphics[width=\textwidth]{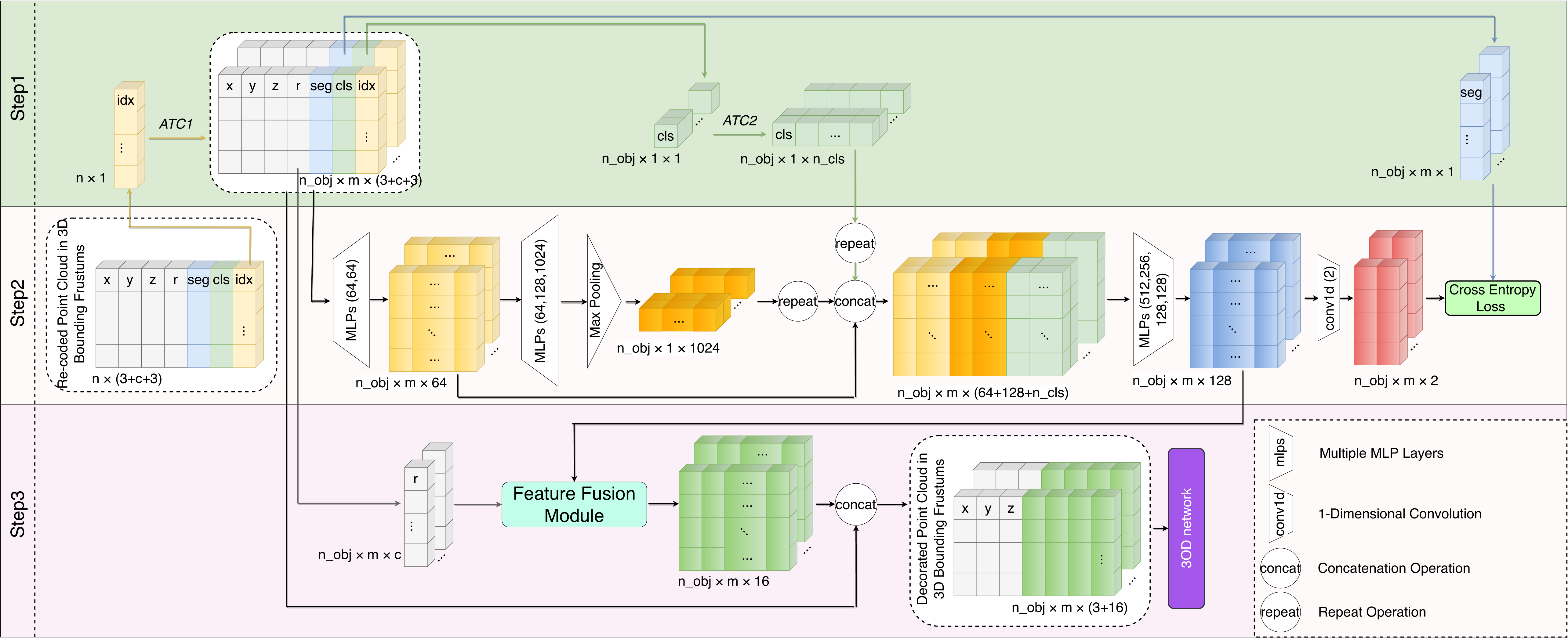}
    \caption{\textbf{Seen module.} It inputs the disordered points with three additional dimensions in the recoded \textit{frustum point clouds} and outputs the decorated points with 16-dimensional semantic features.}
    \label{fig:Seen module}
\end{figure*}

The recoding module point-wisely appends the \textit{frustum point clouds} with  \textit{seg\_label}, \textit{cls\_label} and \textit{index\_label} in each frame. \textit{seg\_label} is the mask information (1 or 0 for foreground or background points), which is obtained by determining whether the point is \textit{inside} ground truth 3D bounding boxes $GT\_3D=\left\{(x,y,z,h,w,l,ry)\right\}_{i=0}^{K-1}$ or not, where $x,y,z,h,w,l,ry$ represent the center coordinates, length, width, height and heading angle of the 3D bounding box, respectively and $K$ represents the number for ground truth 3D bounding boxes in a frame.
\textit{cls\_label} is the category of 2D bounding box to which each point belongs. For the KITTI dataset, \textit{cls\_label} uses 0, 1, 2 and -1 to represent \textit{Car}, \textit{Pedestrian}, \textit{Cyclist} and \textit{background} respectively. 
\textit{index\_label} is the number of 2D bounding box to which each point belongs in the current frame. The above label information is used for the subsequent seen module to perform the 3IS task.

Similar to PointPainting \cite{PointPainting}, the offline operation has some advantages. First, appending additional label information at each point not only does not require changing the complex data loading and processing pipeline of the 3OD network, but also does not add the extra computational load to the 3OD network. More importantly, this offline operation is not affected by any data augmentation of both LiDAR points and images. And there is no problem that the pixels or convolution features of image are difficultly aligned with LiDAR points due to some data augmentation solutions, such as folding, clipping and scaling of the image, rotation of the point cloud and GT-Paste \cite{SECOND}.

\begin{figure}[ht]
    \centering
    \includegraphics[width=0.46\textwidth]{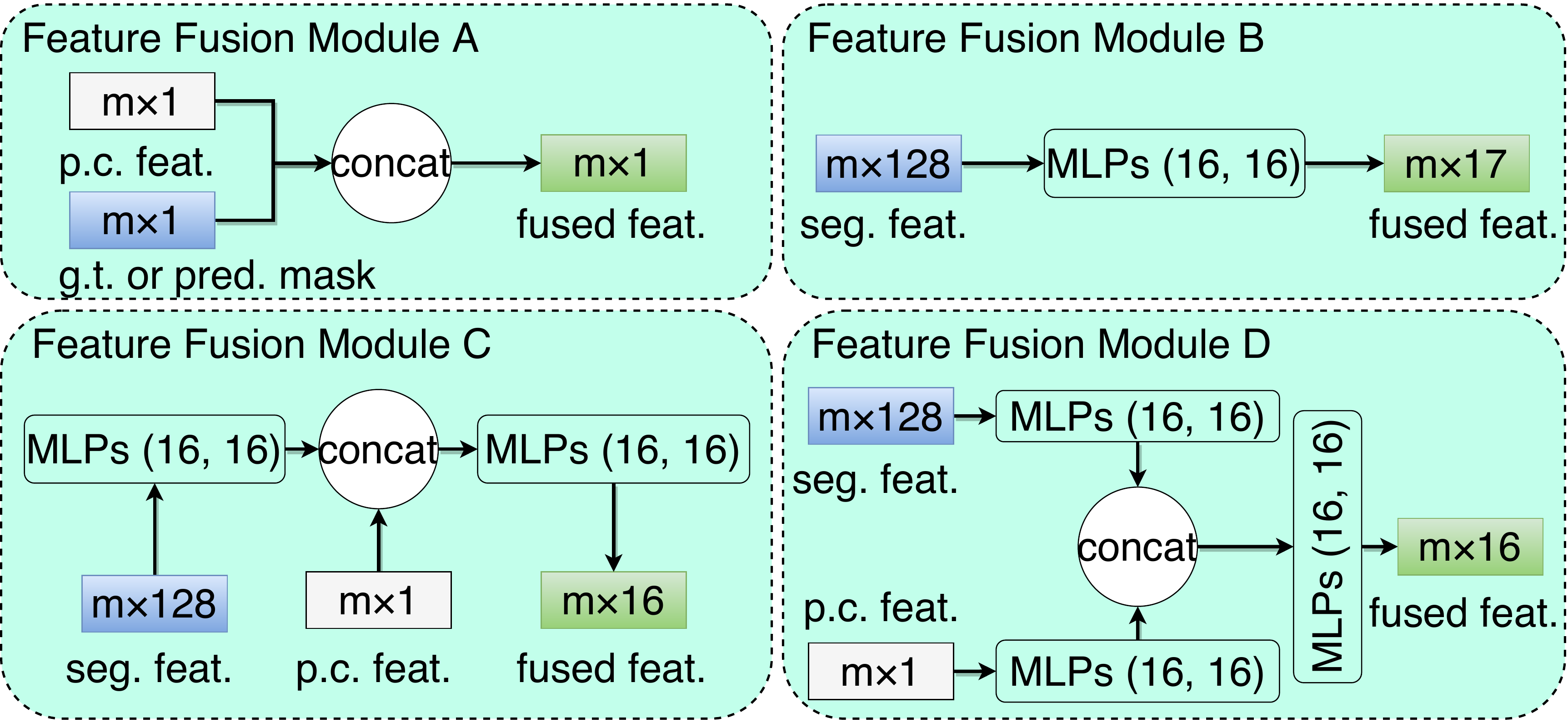}
    \caption{\textbf{Fusion fusion module.} The feature fusion modules A, B, C and D show four feature fusion methods respectively. ``g.t. or pred. feat.'' denotes the ground truth or predicted mask information. ``seg. feat.'' denotes segmentation features. ``p.c. feat.'' denotes features of point cloud itself.  ``fused feat.'' denotes point-wise semantic features used to decorate the point cloud.}
    \label{fig:feature fusion module}
\end{figure}

\subsection{Seen Module}
\textit{Assuming that only the foreground points in the point cloud are fed into the 3OD network, will the detection results be significantly improved?} The answer is definitely Yes. 
Further, if the point cloud has additional mask information (0 or 1 for the foreground or background point), \textit{will the results of the 3OD network also achieve a significant improvement without adding additional computational overhead?}  
To verify the suspicion, we use ground truth 3D bounding boxes to generate the point-wise mask information for the point cloud, which is fed into 3DSSD. As illustrated in the $1^{st}$ and $2^{nd}$ rows of Table~\ref{tab5}, the mask information is of great use for the 3OD network. 
However, the segmentation results of existing lightweight 3IS networks are not particularly accurate, especially for small objects. Therefore, for the detection task, the high-dimensional segmentation features output from the 3IS network are more suitable as additional features for point clouds than the mask information of each point.

Similar to existing point-level fusion approaches, our proposed seen module is designed to obtain point-wise representative semantic features to decorate the point cloud for better performance of any 3OD network. Please note that our method directly acquires point-wise features without the point-to-pixel matching process, thus it can well avoid the inaccurate alignment of features.
As shown in Fig.~\ref{fig:Seen module}, the seen module consists of three steps.

In the first step, as \textit{ACT1} in Fig.~\ref{fig:Seen module}, the disordered point cloud is first divided into multiple \textit{frustum point clouds} based on \textit{index\_label}, then these points in each frustum point cloud are resampled to 2048. After that, as \textit{ACT2} in Fig.~\ref{fig:Seen module}, the \textit{one hot vector} is generated by \textit{cls\_label} for the 3IS task. For the KITTI dataset, \textit{one hot vector} represents \textit{Car}, \textit{Pedestrian}, \textit{Cyclist} by 100, 010 and 001. Besides, the \textit{seg\_label} is used to compute \textit{Cross Entropy Loss} for the 3IS task in the second step.

In the second step, in order to reduce the additional computational overhead, multiple \textit{frustum point clouds} are stacked together along the batch dimension to form a new tensor, which is fed into a 3IS network to obtain high-dimensional segmentation features $F_1$. Notably, the 3IS network in this module is consistent with \textit{3D Instance Segmentation PointNet V1} of F-PointNet \cite{F-PointNet}. However, the predicted background points are not removed. This is because the segmentation of some objects may be incorrect, especially for small objects with few points.

In the third step, in order to decorate point clouds with more representative semantic features, we fuse the acquired high-dimensional segmentation features $F_1$ with features of the point cloud itself $F_2$, such as reflection intensity $r$ for KITTI and RGB values for SUN-RGBD. 
As shown in Fig.~\ref{fig:feature fusion module}, four different fusion modules are designed in order to investigate the effect of mask information, segmentation features and the features of the point cloud itself on the 3OD network. Through experiments (as shown in Table~\ref{tab5}), the result of ``feature fusion module D"  is the best. The ``feature fusion module D" is formulated as 
\begin{equation}
  F_3 = MLPs(Concat(MLPs(F_1), MLPs(F_2)))
\end{equation}
where $MLPs$ represents multiple MLP layers, $Concat$ represents concatenation operation, the features $F_3$ are point-wise representative semantic features for decorating LiDAR points.

The point cloud in Fig.~\ref{fig:Seen module} is an example of the KITTI dataset, in which each point is represented by $x$, $y$, $z$ coordinates and reflection intensity $r$ respectively; \textit{n}, \textit{m}, \textit{c}, \textit{n\_obj} and \textit{n\_cls} are respectively expressed as the number of points in the point cloud, the number of sampled points for each object, the number of features of the point cloud itself, the number of objects in the point cloud and the number of categories (without background). \textit{m} is 2048 for both the KITTI and SUN-RGBD datasets.

\begin{table*}[ht]
\caption{Results on the $Car$ category of the KITTI \textit{test} set. 3DSSD and PoinRCNN are used as baselines. The evaluation metrics \textit{3D Detection}, \textit{Bird's Eye View} and \textit{Orientation} Average Precision are calculated on 40 recall points. The best results are in bold. "L" and "C" indicate the LiDAR and Camera, respectively.}
\label{tab1}
\centering
\scalebox{1}{
\begin{tabular}{c|c|c|ccc|ccc|ccc}
\hline
\multirow{2}{*}{Method} & \multirow{2}{*}{year} & \multirow{2}{*}{Modality} & \multicolumn{3}{c|}{3D Detection} & \multicolumn{3}{c|}{Bird's Eye View} & \multicolumn{3}{c}{Orientation} \\ \cline{4-12} 
                        &&                           & Easy     & Moderate     & Hard    & Easy      & Moderate      & Hard     & Easy    & Moderate    & Hard    \\ \hline
MV3D \cite{MV3D}        & 2017 & L+C                       & 74.97          & 63.63          & 54.00    & 86.62  & 78.93  & 69.80 & -  & -  & -     \\
F-PointNet \cite{F-PointNet}  & 2018 & L+C                       & 82.19          & 69.79          & 60.59    & 91.17  &	84.67  & 74.77 & -  & -  & -     \\
F-Convnet \cite{F-ConvNet}  & 2019 & L+C                       & 87.36          & 76.39          & 66.69    & 91.51  &	85.84  & 76.11 & 95.81  & 91.98  & 79.83     \\
AVOD-FPN \cite{AVOD-FPN}    & 2018 & L+C                       & 83.07          & 71.76          & 65.73    & 90.99  & 84.82  & 79.62 & 94.65 & 88.61 & 83.71 \\
3D-CVF \cite{3D-CVF}     & 2020 & L+C                       & 89.20          & 80.05          & 73.11    & 93.52  & 89.56  & 82.45 & 40.44 & 39.79 & 36.10 \\
CAT-Det \cite{CAT-Det}     & 2022 & L+C                       & 89.87          & 81.32          & 76.68    & 92.59  & 90.07  & 85.82 & 95.95 & \textbf{94.57} & \textbf{91.88} \\ \hline
VoxelNet \cite{VoxelNet}   & 2018 & L                         & 77.47          & 65.11          & 57.73    & -  & -  & - & -  & -  & -      \\
SECOND \cite{SECOND}   & 2018 & L                         & 83.34          & 72.55          & 65.82    & -  & -  & - & -  & -  & -      \\
PointPillars \cite{PointPillars} & 2019 & L                         & 82.58          & 74.31          & 68.99    & 90.07  & 86.56  & 82.81 & 93.84 & 90.70 & 87.47 \\
F-PointRCNN \cite{F-PointRCNN} & 2019 & L                         & 84.28          & 75.73          & 67.39    & -  & -  & - & -  & -  & -      \\
STD \cite{STD}        & 2019 & L                         & 87.95          & 79.71          & 75.09    & 94.74  & 89.19  & \textbf{86.42} & -  & -  & -      \\
PointPainting \cite{PointPainting} & 2020 & L                         & 82.11          & 71.70         & 67.08    & 92.45  & 88.11  & 83.36 & 98.36 & 92.43 & 89.49 \\
SA-SSD \cite{SA-3DSSD}    & 2020 & L                         & 88.75          & 79.79          & 74.16    & 95.03  & \textbf{91.03}  & 85.96 & -  & -  & -      \\
TANet \cite{TANet}      & 2020 & L                         & 83.81          & 75.38          & 67.66    & 91.58  & 86.54  & 81.19 & 93.52 & 90.11 & 84.61 \\
PV-RCNN \cite{pv-rcnn}     & 2020 & L                         & 90.25 & 81.43          & 76.82    & 94.98  & 90.65  & 86.14 & 98.15 & \textbf{94.57} & 91.85 \\
Part-A2 \cite{Part-A2} & 2020 & L                        & 87.81          & 78.49          & 73.51    & 91.70  & 87.79  & 84.61 & 95.00 & 91.73 & 88.86 \\
CIA-SSD \cite{CIA-SSD}      & 2021 & L                         & 89.59          & 80.28          & 72.87    & 93.74  & 89.84  & 82.39 & 96.65 & 93.34 & 85.76 \\
SVGA-Net \cite{SVGA-Net}   & 2022 & L                        & 87.33          & 80.47          & 75.91    & 92.07  & 89.88  & 85.59 & 96.02 & 94.45 & 91.54 \\
IA-SSD \cite{IA-SSA}     & 2022 & L                         & 88.87          & 80.32          & 75.10    & 93.14   & 89.48  & 84.42 & 96.23 & 93.41 & 88.34 \\ \hline
3DSSD \cite{3DSSD}     & 2020 & L                        & 88.36          & 79.57          & 74.55    & 92.66 & 89.02 & 85.86  & -  & -  & -     \\
3DSSD+PointSee         & - & L+C                         & \textbf{91.00} & \textbf{82.04} & \textbf{77.02} & \textbf{95.31} & 89.67 & 84.55 & \textbf{98.68} & \textbf{95.52} & 90.50      \\
\textit{Delta}         & -  & -                         & \textbf{+2.64} & \textbf{+2.68} & \textbf{+2.47} & \textbf{+2.65} & \textbf{+0.65} & -1.31 & -  & -  & - \\ \hline
PointRCNN \cite{PointRCNN}   & 2019 & L                        & 86.96          & 75.64          & 70.70    & 92.13 & 87.39 & 82.72 & 95.90 & 91.77 & 86.92    \\
PointRCNN+PointSee      & - & L+C                         & 89.95          & 79.51          & 74.49    & 94.81 & 88.96 & 83.96 & 98.11 & 92.64 & 87.63  \\
\textit{Delta}        & - & -                         & \textbf{+2.99} & \textbf{+3.87} & \textbf{+3.79} & \textbf{+2.68} & \textbf{+1.57} & \textbf{+1.25} & \textbf{+2.21} & \textbf{+0.87} & \textbf{+0.71}\\ \hline
\end{tabular}
}
\end{table*}

\begin{table*}[ht]
\caption{Results on the ten categories of the SUN-RGBD \textit{test} set. F-ConvNet is used as the baseline. The evaluation metric is \textit{3D} Average Precision (IoU 0.25). The best results are in bold.}
\label{tab4}
\centering
\scalebox{1}{
\begin{tabular}{c|cccccccccc|c}
\hline
Method     & bathtub     & bed     & bookself     & chair     & desk     & dresser    & night\_stand     & sofa     & table     & toilet     & mAP            \\ \hline
F-ConvNet   & 61.32  & 83.19  & 36.46  & 64.40  & 29.67  & 35.10  & \textbf{58.42} & 66.61  & 53.34  & 86.99   & 57.55          \\
F-ConvNet+PointSee & 63.47  & 84.97 & 32.75  & 63.39  & 32.59 & \textbf{34.12} & 57.87  & 67.76 & \textbf{54.91} & 87.68  & \textbf{58.25}          \\
Delta   & \textbf{+2.15}   & \textbf{+1.78} & -3.71 & -1.01 & \textbf{+2.92} & -0.98 & -0.55  & \textbf{+1.15} & \textbf{+1.57} & \textbf{+0.69} & \textbf{+0.70} \\ \hline

\end{tabular}
}
\end{table*}

\subsection{3D Detection}
The decorated point cloud can be fed to various 3OD networks with the point-based encoder, since PointSee just changes the input dimension of LiDAR points and adds the code of the seen module (Fig.~\ref{fig:Seen module}). It is worth noting that PointSee is not suitable for networks with voxel-based and pillar-based encoders. In order to validate that the proposed PointSee is very effective for various point-based 3OD networks, we conduct experiments on the outdoor KITTI and indoor SUN-RGBD datasets, and use the one-stage 3DSSD, two-stage PointRCNN and frustum-based F-ConvNet as baselines.

\subsection{Training Strategy}
2OD networks often suffer from inaccurate 2D bounding box regression, and PointSee is also sensitive to the output of any 2OD network. Inspired by the data enhancement strategy of 2OD networks, a simple yet effective training strategy is designed to solve this problem in the training and inference stage of the 3OD network. In the training stage, the length and width of the 2D bounding box are randomly increased by 0$\%$ - 10$\%$ to accommodate the excessive size of the 2D bounding box. In the inference stage, to alleviate the problem that the size of 2D bounding boxes are too small, the length and width of the 2D bounding box are also increased by 5$\%$ to enlarge the receptive field of the frustum point cloud. This strategy is proved to be effective in the following experiments. In addition, we discard 2D bounding boxes with confidence lower than 0.1 for KITTI and directly use 2D detection results privoded by F-ConvNet.

\begin{table*}[ht]
\caption{Ablation results of our PointSee on the $Car$  category of the KITTI \textit{val} set. 3DSSD is used as the baseline. The evaluation metric \textit{3D Average Precision} is calculated on 11 recall points and 40 recall points. The ``offline fusion module*'' uses ground truth 2D bounding boxes, thus it does not participate in the comparison of results.}
\label{tab2}
\centering
\scalebox{1}{
\begin{tabular}{ccccc|ccc|ccc}
\hline
\multirow{2}{*}{3DSSD} & \multirow{2}{*}{hidden module *} & \multirow{2}{*}{hiddn module} & \multirow{2}{*}{\begin{tabular}[c]{@{}c@{}}seen module\end{tabular}} & \multirow{2}{*}{training strategy} & \multicolumn{3}{c|}{R11} & \multicolumn{3}{c}{R40} \\ \cline{6-11} & & & & & easy & mod. & hard & \multicolumn{1}{c}{easy} & \multicolumn{1}{c}{mod.} & \multicolumn{1}{c}{hard} \\ \hline
\Checkmark             &             &             &             &             & 88.66          & 79.18          & 78.26          & 91.27          & 83.07          & 81.98 \\
\Checkmark             & \Checkmark  &             &             &             & 89.28          & 86.10          & 79.53          & 92.67          & 86.60          & 84.08 \\ \hline
\Checkmark             &             & \Checkmark  &             &             & 88.90          & 79.01          & 78.27          & 92.02          & 82.93          & 80.33 \\
\Checkmark             &             & \Checkmark  & \Checkmark  &             & 89.16          & 84.95          & 78.71          & 92.29          & 85.29          & 82.69 \\
\Checkmark             &             & \Checkmark  &             & \Checkmark  & 88.74          & 84.78          & 78.48          & 91.77          & 85.14          & 82.42 \\
\Checkmark             &             & \Checkmark  & \Checkmark  & \Checkmark  & \textbf{89.57} & \textbf{85.59} & \textbf{79.10} & \textbf{92.78} & \textbf{86.07} & \textbf{83.24} \\ \hline
\end{tabular}
}
\end{table*}

\begin{table}[ht]
\caption{Ablation results of 3DSSD with our feature fusion modules A, B, C and D (as showm in Fig. \ref{fig:feature fusion module}) on the \textit{Car} category of KITTI val set. The evaluation metric \textit{3D Average Precision} is calculated on 11 recall points and 40 recall points. The $1st$ row does not participate in the comparison of results.}
\label{tab5}
\centering
\scalebox{0.9}{
\begin{tabular}{c|ccc|ccc}
\hline
\multirow{2}{*}{Module}                                                & \multicolumn{3}{c|}{R11} & \multicolumn{3}{c}{R40}  \\ \cline{2-7} 
                                                                       & Easy  & Moderate & Hard  & Easy  & Moderate & Hard  \\ \hline
3DSSD                   & 88.66          & 79.18          & 78.26          & 91.27          & 83.07          & 81.98       \\
\begin{tabular}[c]{@{}c@{}}Module A\\ (g.t. mask)\end{tabular} & 90.57  & 88.85 & 88.48  & 96.14 & 90.49& 88.13      \\ \hline
\begin{tabular}[c]{@{}c@{}}Module A\\ (pred. mask)\end{tabular}    & 89.30 & 84.67    & 78.83 & 92.65 & 85.60    & 82.82 \\ 
Module B                                                               & 89.23      & 79.44         & 78.36      &92.43       & 83.46         & 80.60      \\
Module C                                                               & 89.31 & 84.97    & 78.89 & 92.74 & 85.60    & 82.98 \\ 
Module D                                                               & \textbf{89.57} & \textbf{85.59}    & \textbf{79.10} & \textbf{92.78} & \textbf{86.07}    & \textbf{83.24} \\ \hline
\end{tabular}
}
\end{table}

\section{ Experiments}
\subsection{Dataset}
We validate our proposed PointSee on both the popular outdoor KITTI \cite{kitti} and indoor SUN-RGBD \cite{sunrgbd} datasets. 

\textbf{KITTI.} KITTI is a popular benchmark for 3OD in outdoor scenes. It contains 7481 samples for training and 7518 samples for testing. Each sample consists of a point cloud and am RGB image with nine categories (\textit{Car}, \textit{van}, \textit{Truck}, \textit{Pedestrian}, \textit{Cyclist}, \textit{Person\_sitting}, \textit{Tram}, \textit{Misc}, \textit{DontCare}). Following the commonly applied setting, we divide all training samples into 3712 samples and 3769 samples for train split and val split. For submission to the KITTI test server, we follow the training strategy mentioned in SASA \cite{sasa}, where 80$\%$ training samples are used for training and the rest 20$\%$ training samples are used for validation.

\textbf{SUN-RGBD.} Unlike the KITTI dataset, SUN-RGBD is a 3OD dataset for indoor scenes. It contains 5285 RGB-D images for training and 5050 RGB-D images for testing.

\begin{figure*}[ht]
    \centering
    \includegraphics[width=\textwidth]{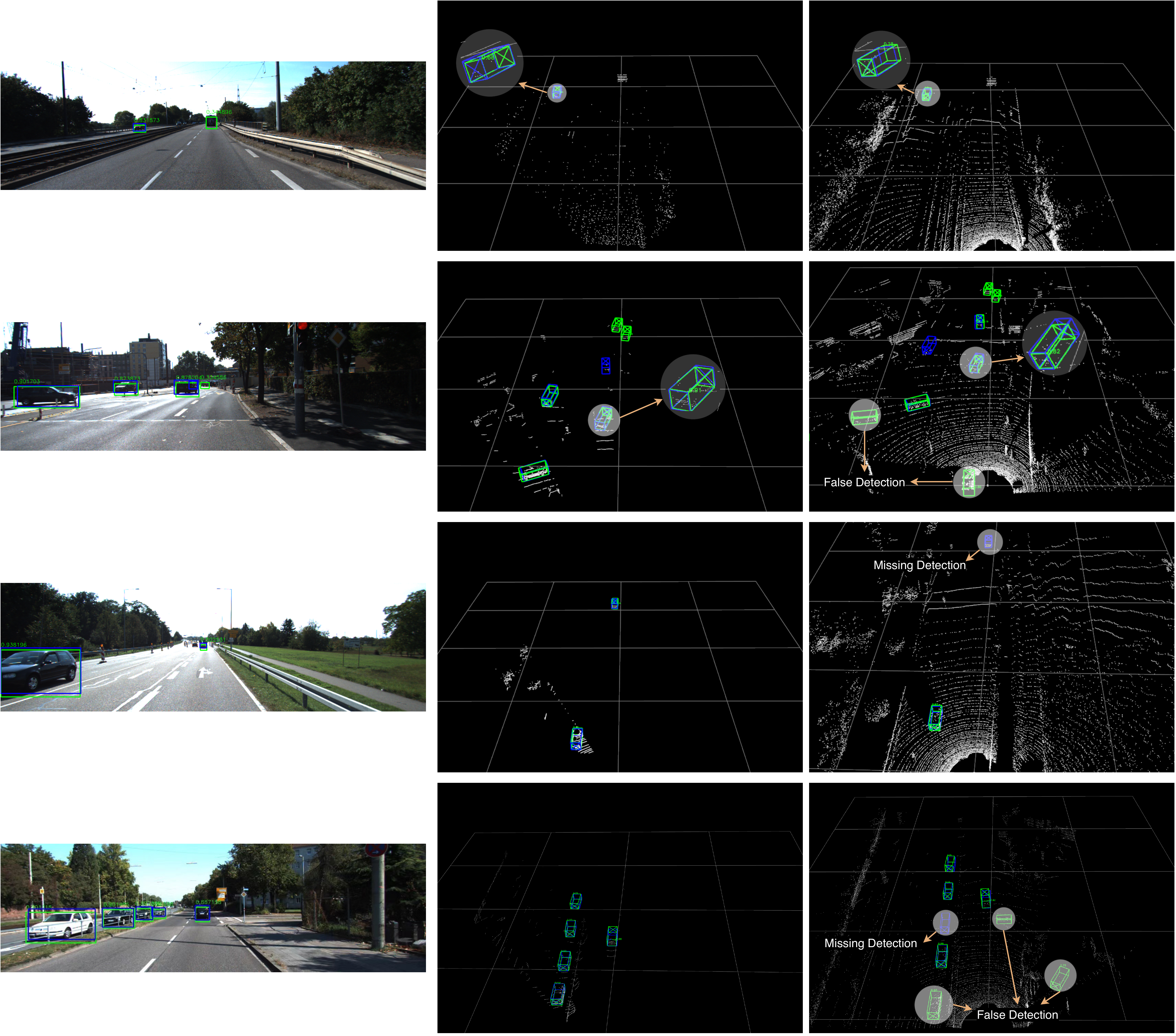}
    \caption{Visualization of detection results of Yolov5 (the $1^{st}$ column), 3DSSD with PointSee (the $2^{nd}$ column) and 3DSSD (the $3^{rd}$ column) on the KITTI val split. The predicted boxes and ground-truth boxes are labeled in green and blue, respectively.}
    \label{fig:results viz}
\end{figure*}

\begin{figure*}[ht]
    \centering
    \includegraphics[width=\textwidth]{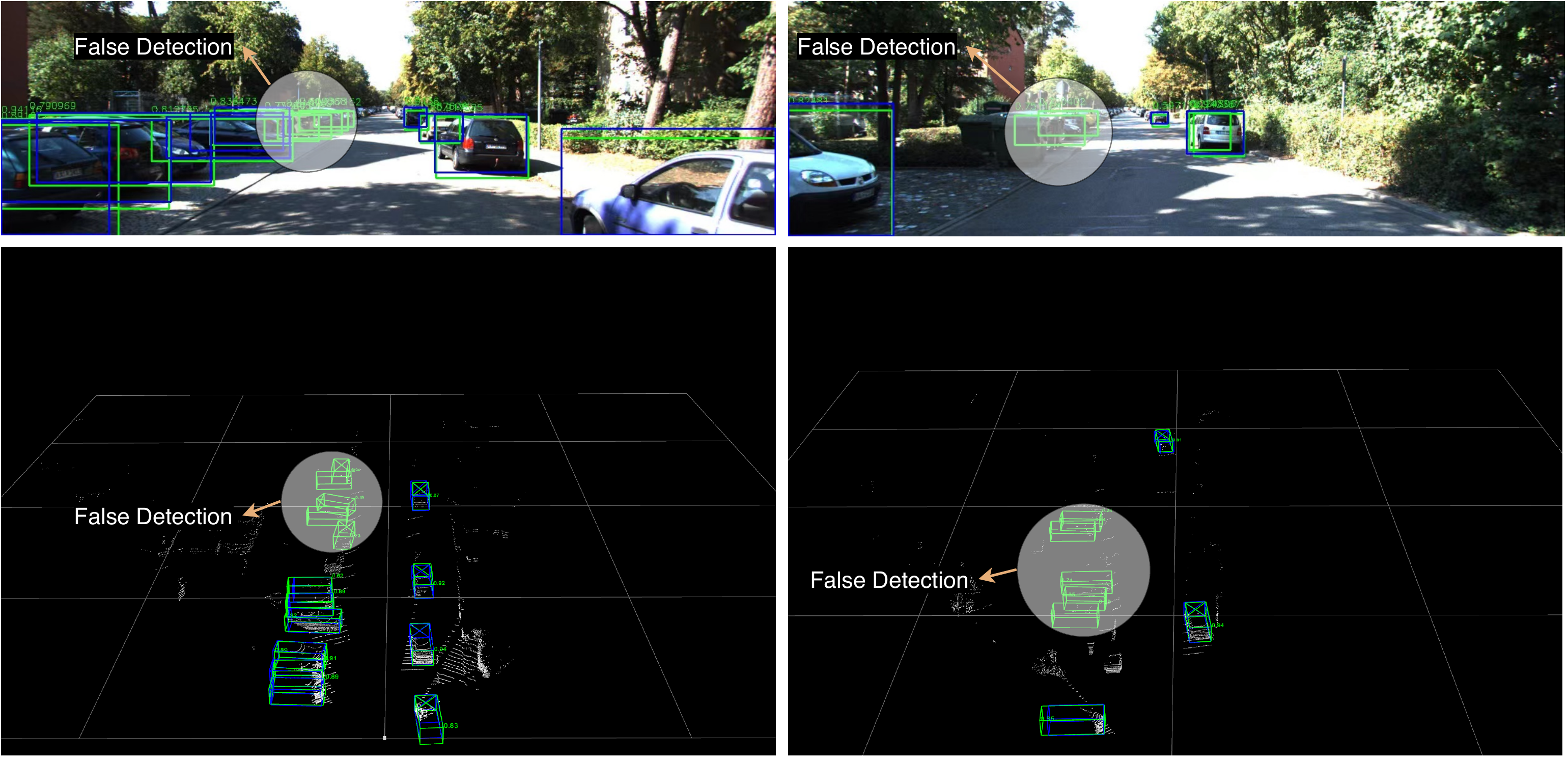}
    \caption{Visualization of failure cases of Yolov5 (the $1^{st}$ row) and 3DSSD with PointSee (the $2^{nd}$ row) on the KITTI val split. The predicted boxes and ground-truth boxes are labeled in green and blue, respectively.}
    \label{fig:results viz false}
\end{figure*}

\begin{table*}[ht]
\caption{Results of 3DSSD and PointRCNN with and without PointSee on the $Car$ category of the KITTI \textit{val} set. The evaluation metric \textit{3D Average Precision} is calculated on 11 recall points and 40 recall points.}
\label{tab3}
\centering
\scalebox{1}{
\begin{tabular}{c|ccc|ccc|c}
\hline
\multirow{2}{*}{Method} & \multicolumn{3}{c|}{R11} & \multicolumn{3}{c|}{R40} & \multirow{2}{*}{Inference Time} \\ \cline{2-7}
                        & Easy    & Moderate    & Hard   & Easy    & Moderate    & Hard   &                      \\ \hline
3DSSD                   & 88.66          & 79.18          & 78.26          & 91.27          & 83.07          & 81.98   & 18.89 ms/frame      \\
3DSSD+PointSee           & \textbf{89.57} & \textbf{85.59} & \textbf{79.10} & \textbf{92.78} & \textbf{86.07} & \textbf{83.24} & 27.21 ms/frame \\
Delta                   & \textbf{+0.91} & \textbf{+6.41} & \textbf{+0.84} & \textbf{+1.51} & \textbf{+3.00} & \textbf{+1.26} & 8.32 ms/frame\\ \hline
PointRCNN               & 89.20          & 78.87          & 78.46          & 91.83          & 82.40          & 80.26    & 40.18 ms/frame     \\
PointRCNN+PointSee       & \textbf{89.48} & \textbf{84.49} & \textbf{78.76} & \textbf{92.87} & \textbf{84.92} & \textbf{80.98} & 44.90 ms/frame\\
Delta                   & \textbf{+0.28} & \textbf{+5.72} & \textbf{+0.30} & \textbf{+1.04} & \textbf{+2.52}  & \textbf{+0.72} & 4.72 ms/frame\\ \hline
\end{tabular}
}
\end{table*}

\subsection{Experiment Settings}
To widely verify PointSee, we select three different and representative baselines 3DSSD, PointRCNN and F-ConvNet for evaluation. Our experimental models run on single GTX3090. In addition, we use Yolov5 for 2D detection on the KITTI dataset, and directly used the 2D detection results of the SUN-RGBD dataset provided by F-ConvNet \cite{F-ConvNet}.

\textbf{3DSSD and PointRCNN.} For the KITTI dataset, we train 3DSSD and PointRCNN in an end-to-end manner with the same parameter configuration as SASA \cite{sasa}. In addition, since the hidden module is completed offline, there is no need to care about the data augmentation of the original 3OD network. We only change the number of channels dedicated to input the point cloud and add the code of the seen module.

\textbf{Yolov5.} For the KITTI dataset, we use the largest model \textit{Yolov5x} to train in an end-to-end manner with the SGD optimizer for 100 epochs and use an image size of 1280 pixels in the training and inference stage. In addition, we merge \textit{Car}, \textit{Truck} and \textit{Van} into the \textit{Car} category, and only retain the \textit{Car}, \textit{Pedestrian} and \textit{Cyclist} categories for training and inference.  We follow the original data augmentation strategies, training and inference settings.

\textbf{F-ConvNet.} For the SUN-RGBD dataset, we train F-ConvNet without changing the parameters and use the same 2D detection results as F-ConvNet.

\subsection{Main Results}
We evaluate the effectiveness of our proposed PointSee on both the KITTI and SUN-RGBD datasets. For the KITTI dataset, our evaluation is performed on the point-based 3D detection networks represented by the one-stage 3DSSD and two-stage PointRCNN for outdoor scenes. For the SUN-RGBD dataset, we use F-ConvNet as the baseline to evaluate the effectiveness of PointSee for indoor scenes.

\textbf{Results on the KITTI Dataset.} Table~\ref{tab1} shows the 3D object detection performance on the KITTI test set evaluated on the official server. For the most competitive \textit{Car} category, the 3DSSD with PointSee surpasses all existing 3D detection networks for the easy level of \textit{3D Detection}, \textit{Bird's Eye View} and \textit{Orientation} average precision, and obtains comparable results to state-of-the-art PV-based models (such as PV-RCNN). Comparing with the baseline model 3DSSD, our method boosts the 3D AP by 2.64$\%$, 2.68$\%$, 2.47$\%$ for the three difficulty levels respectively. In addition, our method helps PointRCNN to boost the 3D AP by 2.99$\%$, 3.87$\%$, 3.79$\%$ for the three difficulty levels respectively. These experimental results show that our method is applicable to arbitrary point-based 3OD networks and can significantly improve the performance of 3OD.

%It is worth noting that our method obtains significant improvements on the easy, moderate and hard levels of \textit{3D Detection}, \textit{Bird's Eye View} and \textit{Orientation} average precision, showing that our proposed PointSee can help 3OD networks to focus on regions where objects of interest may exist and obtain more representative features, which facilitates to retain key points of almost invisible instances (e.g., small or occluded cars) so as to make more robust 3OD.

\textbf{Results on the SUN-RGBD Dataset.} As shown in Table~\ref{tab4}, we use F-ConvNet as the baseline to evaluate PointSee in the SUN-RGBD dataset. PointSee enables F-ConvNet to achieve better performance in six categories, and boosts the mAP by 0.7$\%$. This further validates the flexibility and scalability of our proposed PointSee.

\subsection{Ablation Study}
We conduct ablation studies to validate each part of PointSee. The results provided in Table~\ref{tab2} and Table~\ref{tab5} are trained on the KITTI \textit{train} split and evaluated on the KITTI \textit{val} split of the \textit{Car} category.

\textbf{Effects of Hidden Module.} The $1^{st}$ and $2^{nd}$ rows of Table~\ref{tab2} verify the effectiveness of \textit{frustum point clouds}, which helps the 3OD network to focus more on the regions of point clouds where objects of interest may exist. Note that the $2^{nd}$ column ``offline fusion module*'' in Table~\ref{tab2} uses ground truth 2D bounding boxes, while the $3^{rd}$ column ``offline fusion module'' in Table ~\ref{tab2} uses the output of the 2OD network Yolov5. From the experimental results of the $1^{st}$ and $3^{rd}$ rows of Table ~\ref{tab2}, using the 2D bounding boxes output from Yolov5 for the seen module will degrade the performance of 3DSSD for the moderate and hard levels. This is because 2D bounding boxes output from Yolov5 may be too small or too large, resulting in some foreground points of objects being erroneously discarded or some background points being mistakenly retained. 

\textbf{Effects of Seen Module.} The $3^{rd}$ and $4^{th}$ rows of Table ~\ref{tab2} compare the 3D detection performance of 3DSSD with and without the seen module. This module significantly boosts the 3D AP for the easy, moderate and hard levels on 11 and 40 recall points, especial by 5.94$\%$ for the moderate level on 11 recall points. Moreover, as shown in the $5^{rd}$ and $6^{th}$ rows of Table ~\ref{tab2}, the results also validate the effectiveness of the module in helping point clouds to extend more representative semantic features, which is very helpful in improving the performance of 3OD networks.

\textbf{Effects of Feature Fusion Module.} Fig.~\ref{fig:feature fusion module} shows four different feature fusion modules, and the experimental results are shown in Table \ref{tab5}. Module A (g.t. mask) verifies our suspicion that the additional mask information for the point cloud can help the 3OD network to obtain a significant performance improvement. Further, Module A (pred. mask) decorates the point cloud with the predicted mask information, and the results are much different than those of module A (g.t. mask). This is mainly caused by two reasons: firstly, the lightweight 3IS network is not very effective for small objects, and secondly, the 2OD network may have the problem of missing detection. To solve the problem, we use the high-dimensional segmentation features output from the 3IS network instead of the mask information. The $5^{th}$ (Module C) and $4^{th}$ (Module B) rows of Table \ref{tab5} show the effectiveness of the segmentation features and the features of the point cloud itself. The $6^{th}$ (Module D) row of Table \ref{tab5} verifies that the semantic features of the point cloud itself extracted after the 2-layer MLPs can be better fused with the segmentation features to further assist the 3OD network.

\textbf{Effects of Training Strategy.} The $4^{th}$ and $6^{th}$ rows of Table~\ref{tab2} verify that our proposed training strategy can effectively increase the robustness of the 3OD network and reduce their sensitivity to 2D bounding box regression (too large or too small) of the 2OD network. In addition, it is worth noting that the $6^{th}$ row of Table ~\ref{tab2} obtains comparable results to the $2^{nd}$ row of Table ~\ref{tab2}, which further demonstrates the superiority of our PointSee.

\subsection{Compatibility Study}
Our PointSee is an easy-to-plug-in design and can be applied to various point-based 3OD networks. As shown in Table~\ref{tab3}, PointSee obtains notable enhancement on the one-stage model 3DSSD and the two-stage model PointRCNN.

\textbf{Result on the KITTI Dataset.} The backbone of 3DSSD and PointRCNN have the SA layers, which extract a small number of key points for object localization and 3D bounding box regression. In these SA layers, some points of small objects or occluded objects are easily discarded by mistake, which leads to missed detection. PointSee is helpful to avoid this problem. As shown in Table ~\ref{tab3}, it improves the detection performance on 3DSSD and PointRCNN for all difficulty levels. For harder instances with fewer points, PointSee can better focus on these points, and acquire more representative semantic features to aid subsequent 3OD.

\textbf{Inference Speed.} As shown in the $8^{th}$ column of Table ~\ref{tab3}, our PointSee adds 8.32 ms and 4.72 ms of additional time consuming per frame for 3DSSD and PointRCNN, respectively. This time consuming is worthwhile because PointSee brings significant performance improvement to the 3OD network. 

\begin{figure}[ht]
    \centering
    \includegraphics[width=0.5\textwidth]{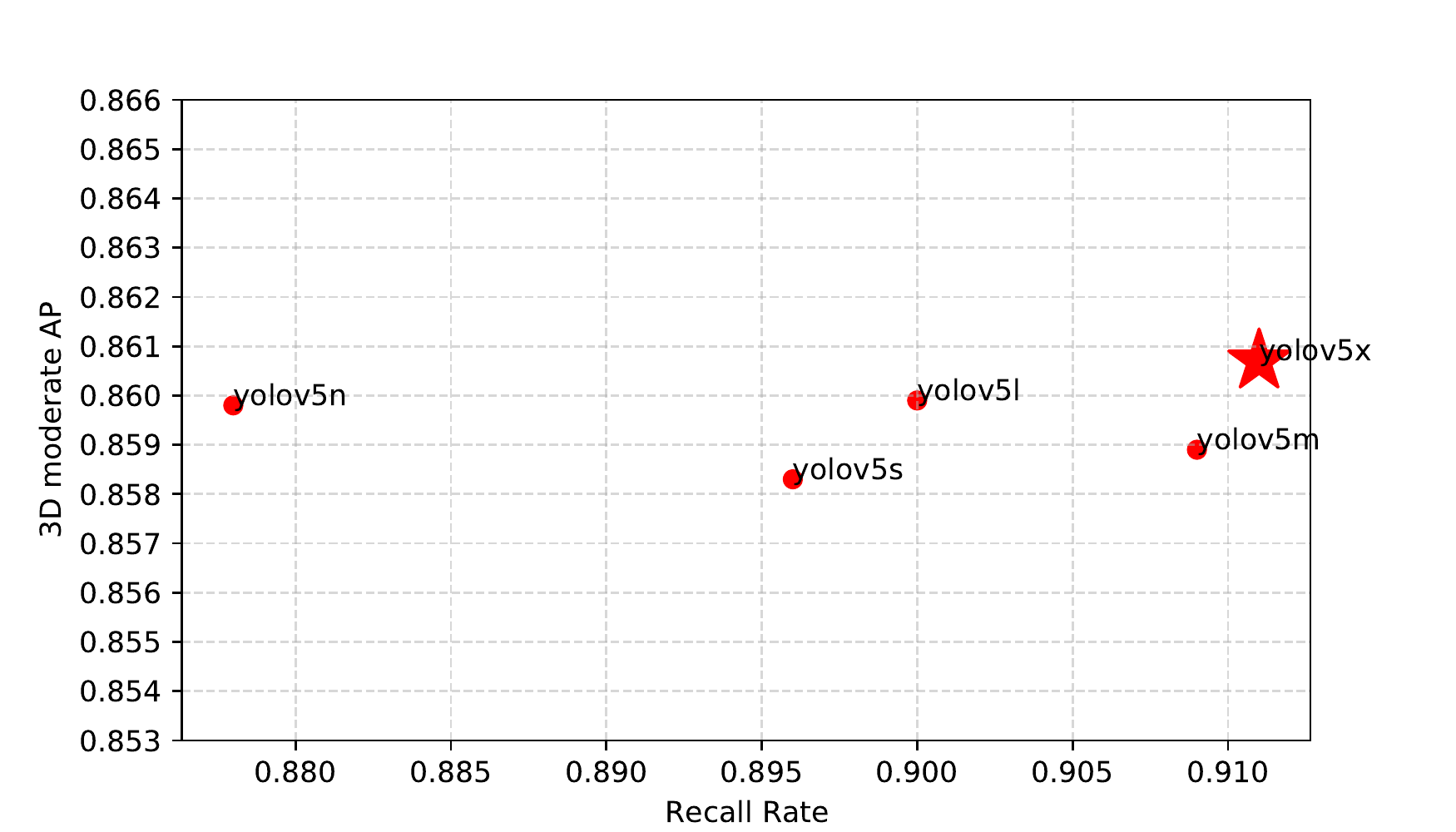}
    \caption{PointSee depends on the quality of the 2OD network on the KITTI val split. The performance of 3DSSD with PointSee, as measured by \textit{3D Average Precision} calculated on 40 recall points, is compared with respect to the quality of the 2OD network used in the hidden module, as measured by \textit{Recall Rate}.}
    \label{fig:results line picture}
\end{figure}

\textbf{Dependency on the 2OD network.} We utilize five kinds of Yolov5 models to evaluate the effect of different 2OD networks on PointSee. As illustrated in Fig.~\ref{fig:results line picture}, the models from small to large are \textit{Yolov5n}, \textit{Yolov5s}, \textit{Yolov5m}, \textit{Yolov5l} and \textit{Yolov5x} respectively, and their recall also increases with the size of the models. However, we find that \textit{Recall Rate} of the 2OD network does not significantly affect the performance of PointSee. Concretely, the best results (moderate AP: 86.07$\%$) differ from the worst results (moderate AP: 85.83$\%$) by only 0.24$\%$. Even the smallest model \textit{Yolov5n} can enable PointSee to obtain the second best result (moderate AP: 85.98$\%$). The above experimental results verify that PointSee is of good robustness for different 2OD networks.

\textbf{Quantitative Analysis.} In Fig.~\ref{fig:results viz}, we visualize representative detection results of Yolov5 (the $1^{st}$ column), 3DSSD with PointSee (the $2^{nd}$ column) and 3DSSD (the $3^{rd}$ column) on the KITTI val split. PointSee assists the 3OD network to significantly reduce false detection (see the $2nd$ row of Fig.~\ref{fig:results viz}) and missing detection (see the $3rd$ row of Fig.~\ref{fig:results viz}), which benefits from the fact that the \textit{frustum point clouds} reduces the field of view of the 3OD network and only focuses on the area where there may be objects. In addition, as illustrated in the $1^{st}$ and $2^{nd}$ rows of Fig.~\ref{fig:results viz}, PointSee also enables the 3OD network to obtain more representational semantic features for harder instances (e.g., small or occluded objects), which can facilitate the detection of these harder instances and generate more accurate 3D bounding boxes.

\subsection{Limitations}
%Although Fig. \ref{fig:results line picture} demonstrates that our PointSee is robust to the 2OD network, the performance of PointSee is still limited due to the fact that the 2OD detector fails in detecting small or occluded objects and dense non-rigid objects such as pedestrians. 
For objects that fail to be detected by the 2OD network, PointSee also ``blinds" the 3OD network. And for objects that are incorrectly detected by the 2OD network, PointSee instead enhances the features of the point cloud in this region, which leads to false detection. 
%Besides, although the hidden module has many advantages, it splits the 2OD network and 3OD network into two tasks, resulting in a gap between 2D domains and 3D domains. 

\section{Conclusion}
LiDAR and camera, as two different sensors, are widely equipped on autonomous driving vehicles. Images from the camera can help understand point clouds from the LiDAR for better 3D applications. In this paper, we propose an effective point semantic feature enhancement module, called PointSee. PointSee is lightweight and can be flexibly embedded into arbitrary 3OD networks to improve their detection performance.
Experimental results on both the outdoor KITTI and indoor SUN-RGBD datasets validate that PointSee assists the 3OD networks to focus more on regions where objects may be present in the point cloud, and the obtained  representative semantic features can guide the point-based backbone networks to better model potential objects. In the future, we will investigate how to use the lightweight 2OD network trained jointly with arbitrary 3OD networks to solve the occasional blindness of the 3OD network.

\bibliographystyle{IEEEtran}
\bibliography{bare_jrnl_compsoc}

\end{document}